# Using Person Embedding to Enrich Features and Data Augmentation for Classification

Ahmet Tuğrul Bayrak
*Research and Development Center*
*Huawei Turkey*
İstanbul, Turkey
ahmet.tugrul.bayrak1@huawei.com

*Abstract*—Today, machine learning is applied in almost any field. In machine learning, where there are numerous methods, classification is one of the most basic and crucial ones. Various problems can be solved by classification. The feature selection for model setup is extremely important, and producing new features via feature engineering also has a vital place in the success of the model. In our study, fraud detection classification models are built on a labeled and imbalanced dataset as a case-study. Although it is a natural language processing method, a customer space has been created with word embedding, which has been used in different areas, especially for recommender systems. The customer vectors in the created space are fed to the classification model as a feature. Moreover, to increase the number of positive labels, rows with similar characteristics are re-labeled as positive by using customer similarity determined by embedding. The model in which embedding methods are included in the classification, which provides a better representation of customers, has been compared with other models. Considering the results, it is observed that the customer embedding method had a positive effect on the success of the classification models.

## I. INTRODUCTION

Classification is one of the most popular and widely used methods of machine learning applications. The usage of classification models can be observed in various fields such as; healthcare, e-commerce, banking, agriculture, etc. The data and labels used in the classification models are extremely important for the success of the models. However, the way the data is used and the features produced are also very effective as well. When data and features are of higher quality and more representative, model success also increases from there [1].

One of the most critical problems in classification problems is fraud detection Many malicious transactions can be prevented thanks to fraud detection, in which harmful transactions made by individuals are detected. However, detecting fraudulent actions is a difficult task in the virtual environment, as both legal and fraudulent transactions are observed to have similar behavior tendencies [2]. This is why fraud detection involves a difficult process of monitoring users' behavior to predict and detect or prevent undesirable behavior.

When the studies in the literature are examined, it is understood that credit card fraud detection algorithms are mostly based on data mining. For example, in credit card fraud, suspicious transactions can be identified by dividing them into two classes legitimate (legal) and fraudulent transactions [3], [4]. In addition, algorithms such as decision tree, artificial neural network, support vector machine (SVM), genetic algorithms, naive Bayes algorithm, fuzzy logic and logistic regression are frequently applied in fraud detection processes [5]–[8]. Due to the increase in credit card fraud in recent years, it is observed that there is an increase in the studies on this subject in the literature. In a study in this area, [9], the performances of logistic regression and naive Bayes algorithms are compared. The analysis results point to several cases where the performance of logistic regression is lower than that of naive Bayes, but this is reported to occur especially in small datasets. In another study, three different classification methods (decision tree, neural network, and logistic regression) are tested for their applicability in fraud detection [10]. The results revealed that neural networks and logistic regression approaches outperform the decision tree in solving the related problem. Another study investigating the detection of credit card fraud using decision trees and support vector machines emphasizes that the decision tree approach outperforms the SVM approach in solving the investigated problem [11]. While testing the training data, it is stated that the SVM-based model is close to the success of the decision tree-based model. However, this algorithm is reported to be insufficient in real-time fraud detection. Another related study compares the performance of SVM, random forest and logistic regression in detecting credit card fraud [12]. This study shows that logistic regression has similar performance to random forest algorithms with different low sampling levels, while SVM gives relatively good results despite a lower fraud rate in training data. It has also been noted that logistic regression often performs remarkably, exceeding the performance of the SVM model. In another study, time series analysis is applied to credit card fraud detection [13]

Independently from this topic, there is a phenomenon in recent years called word embedding [14]. Especially starting with word2Vec [15], word embeddings have been one of the hottest topics in natural language processing (NLP). Various embedding methods have been created. Apart from the applications in NLP, word embedding methods have been started to use to embed some other entities like; products, items, people, etc.

In our study, a fraud detection dataset is used. Person embedding is applied to the customers in this labeled dataset, according to the categories they are interested in. In this way, an embedding vector of the specified size was created for each customer and these vectors are added as a feature to the dataset. It has been observed that the success of classification has increased slightly when adding embedding vectors, which also show the similarity of customers according to the product categories they buy, as a feature [16].

## II. METHOD

In this section; the features of data and the preprocessing steps applied to data will be explained. Afterward, the details of the word embedding methods and their usage to create customer vectors will be shared. Finally, the details of the classification model will be provided.

### A. Data and Preprocessing

For the study, an open dataset being available at Kaggle is used[1]. The dataset contains credit card transitions with a label indicating that the transaction is fraudulent or not. The data consists of 1852384 rows. To prepare the features some preprocessing steps are applied. *category* feature containing 14 unique values is one-hot encoded. Using the transaction date, the day of the transaction is extracted. Besides the time of the transaction is also extracted and assigned to one of the bins based on the hour interval (06-12, 12-18, 18-00, 00-06). After that, gender is transformed into 1 and 0 values. The number of locations per customer is calculated and added as a feature as well. The customers do not have unique customer identification numbers in the dataset. For this reason, to represent a customer, several features are combined. The features combined are *first_name*, *last_name*, *job*, *date_of_birth*, *home_address*. The customer id is created by concatenating and hashing them.

### B. Word Embedding Model

Word embedding is a powerful mechanism. Its basic usage is to represent words in a text corpus. A neural network is run behind and a word-space is created as a result based on the similarity of words created with numerous different dimensions [14]. This word space can be used to find the most similar words for a given one.

Even if word embedding is an NLP topic, it has been applied in many different areas. Instead of words, an item, a customer, an event, or more generally an entity can be embedded after being provided in a valid format [16]. In a case when a group of the product is fed into a word-embedding mechanism, it may recommend similar items, which is also quite popular in recent years [17].

In the study, customers are embedded using their generated customer ids instead of words using the same logic (Figure 1). The customers, who have the items in the same category in the same week, are assumed to be similar and customer vectors are added to the features to find out the answer. It might be considered the same as one-hot encoding of categories. However, the difference is that a neural network is run background which can detect and reason patterns, the time of the transaction is another factor for embedding. The order of customers in a list (corresponding to a sentence in NLP-sense) is formed based on that. That differentiates between using an embedding model and one-hot encoding even if they derive from the same feature.

[1]https://www.kaggle.com/code/rosicky1234/credit-card-transactions-fraud-detection-dataset/data

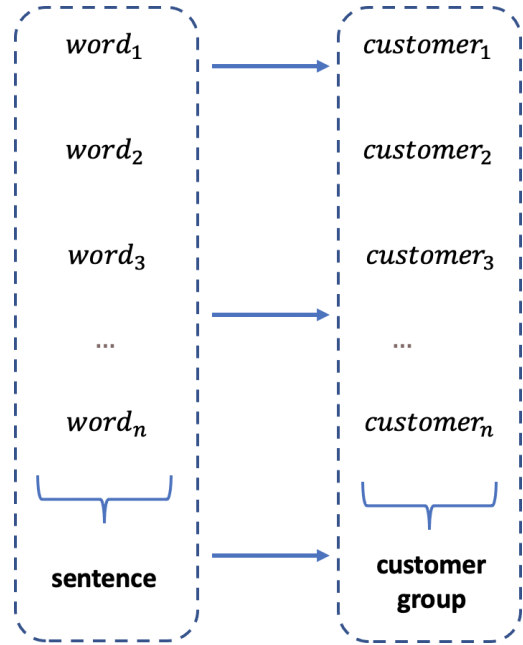

Fig. 1: Sentence - Customer Group

TABLE I: EMBEDDING MODEL PARAMETERS

| Model | Parameters |
| --- | --- |
| embedding size | 20 |
| window size | 40 |
| min word | 5 |
| iter | 100 |

For embedding, one of the best embedding models [18], the FastText model is created and a customer space is established. The embedding model is built using the parameters on Table I.

### C. Classification Models

To be able to see the effect of customer features created with embedding, several classification models are built and compared. After the preprocessing step, the data consists of 1792384 rows. The labels are imbalanced. The number of 1's is 18506 and 0's is 1773878, which represents whether a transaction is fraudulent or not. Before building models, 30% of the data is separated as the test set to evaluate the model's success. After that, the models; decision tree (DT), k-nearest neighbors (KNN), random forest (RF), logistic regression (LR), support vector classifier (SVC), multilayer perceptron (MLP) are built with the features in Table II.

TABLE II: MODEL FEATURES

| Feature | Explanation |
|---|---|
| amount | transaction amount |
| day_of_week | one-hot features for days of week |
| time_interval | time in a day is separated into 4 parts (06-12, 12-18, 18-00, 00-06) |
| customer_embeddings | customer embedding vectors |
| transaction_continent | the continents of transactions occur |
| category | one-hot encodes of categories |
| continent | the continent of the transaction |
| gender | encoded as 1 for female, 0 for male |
| number_of_locations | number of different transactions per customer |

TABLE III: CLASSIFICATION MODEL PARAMETERS

| Models | Parameter Values |
|---|---|
| DT | max_depth=3 |
| KNN | algorithm='auto', leaf_size=30, metric='minkowski', n_neighbors=14, p=2, weights='distance' |
| RF | bootstrap=False, max_depth=50, max_features=10, min_samples_leaf=5, min_samples_split=8, n_estimators=100, random_state=10 |
| LR | C=0.001, penalty='l2' |
| MLP | alpha=0.05, hidden_layer_sizes=(100, ), dropout=0.1, activation='relu', solver='adam', max_iter=100 |
| SVC | C=10, gamma=0.01, kernel='rbf', random_state=10 |

The parameters which are used in building the models can be seen in Table III. The parameters are optimized via grid search.

### D. Data Augmentation via Customer Embedding Similarity

The positive labels in the dataset are relatively small. Because of the imbalanced dataset, models may favor the dominant label [19]. Several data augmentation methods exist. For the study, probably the most popular method SMOTE [20], is applied as a baseline.

Apart from that, customer similarities are calculated using customer vectors. After that, the customers with the label 0 but quite similar (higher than 0.95) to a customer with the label 1, are re-labeled as 1. At the end of this step, the number of positive labels increases from 18506 to 32838.

### III. EXPERIMENTS

For the study, several models are built and their results are compared. The first model built is the base model. The features of the model can be seen in Table II and the feature importance of the base model is in Figure 2. The customer embeddings are formed from categories, therefor another model is built with one-hot encoded category features. After these models, another model is created with customer embeddings. The dimensions of embeddings are used as features.

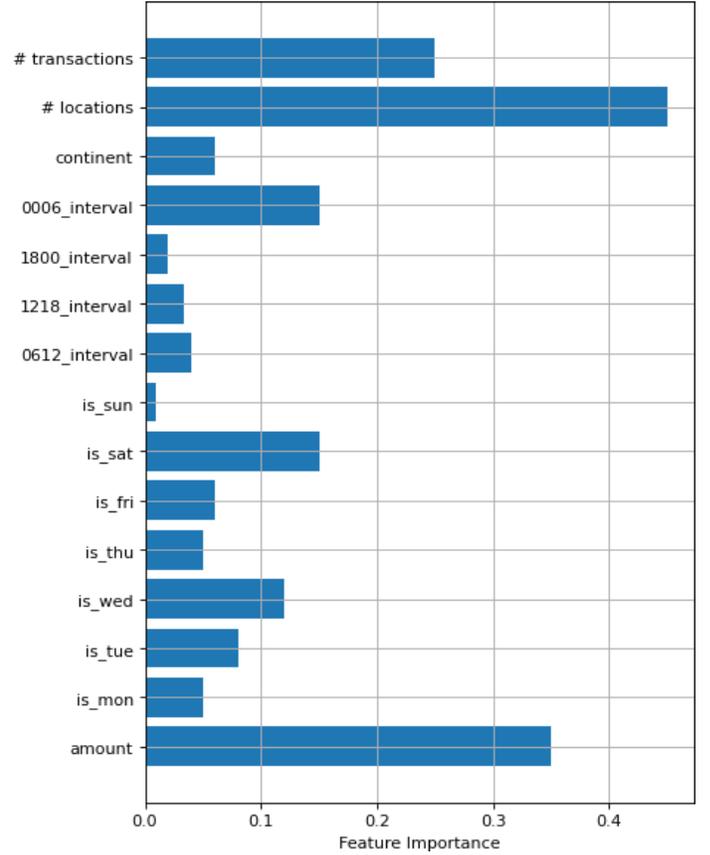

Fig. 2: Feature importance of the base model

As it was mentioned earlier, the labels on the dataset are imbalanced (18506 positives, 1773878 negative labels). Customer embedding is applied to increase the number of positive labels as well. Since there is a customer space, customer similarity can be measured based on that. By using customer similarity, the rows containing the customers who are a close minimum of 0.95 to the ones with label 1 are re-labeled as 1. After that process, the number of positive labels in the dataset increases to 32838.

To compare the data augmentation with customer similarity, SMOTE is also applied. The number of positive labels rises to 35772.

### IV. OBSERVATIONS

In this section, we compare the results of the models. The results are on Table IV. There are 20 different models built. Model group 1 has the baseline models with one-hot encodes of categories. The second model groups are the ones built with customers embedding as features. The models in the third group have baseline models in addition to augmented labels with SMOTE. And finally, the models in group 4 are built with customer embeddings and label augmentation with customer similarity.

TABLE IV: RESULTS WITH DEFAULT MODEL

| Groups | Method | Precision | Recall | F1 Score |
|---|---|---|---|---|
| Group 1 | DT | 0.612 | 0.643 | 0.6271 |
| | RF | 0.665 | 0.671 | 0.6679 |
| | LR | 0.652 | 0.644 | 0.6479 |
| | KNN | 0.659 | 0.658 | 0.6585 |
| | MLP | 0.680 | 0.695 | 0.6874 |
| | SVC | 0.691 | 0.683 | 0.6869 |
| Group 2 | DT | 0.634 | 0.662 | 0.6476 |
| | RF | 0.679 | 0.688 | 0.6834 |
| | LR | 0.672 | 0.661 | 0.6664 |
| | KNN | 0.669 | 0.670 | 0.6695 |
| | MLP | 0.696 | 0.711 | 0.7034 |
| | SVC | 0.717 | 0.702 | 0.7094 |
| Group 3 | DT | 0.622 | 0.639 | 0.6303 |
| | RF | 0.669 | 0.672 | 0.6704 |
| | LR | 0.654 | 0.648 | 0.6509 |
| | KNN | 0.663 | 0.671 | 0.6669 |
| | MLP | 0.682 | 0.691 | 0.6864 |
| | SVC | 0.694 | 0.687 | 0.6904 |
| Group 4 | DT | 0.643 | 0.674 | 0.6581 |
| | RF | 0.691 | 0.703 | 0.6969 |
| | LR | 0.684 | 0.670 | 0.6769 |
| | KNN | 0.688 | 0.689 | 0.6885 |
| | MLP | 0.711 | 0.726 | 0.7184 |
| | SVC | 0.719 | 0.716 | 0.7174 |

As it can be observed from the table, using customer embedding as features performs better than one-hot encoded categories. Even if the source of the features is the same, the embedding model considers the order of the customer group instead of whether they just have a category or not. When the effects of label augmentation are examined, the methods using customer similarity are more successful than SMOTE, since it analyzes better the relationship between customers.

## V. CONCLUSION

Fraud detection is a vivid topic for machine learning, especially for credit cards. As a classification problem, numerous different approaches have been proposed and applied. In this study, we add features related to customers with the question, how will customer representation is affected the classification results. For customer representation, customers are embedded using FastText based on the category they have bought. Therefore, a customer space is created with 20 dimensions. The values representing customers are fed as features.

To detect fraudulent transactions, several models are built. To see the performance of customer vectors coming from embedding are added to features and the models are compared. Apart from that, categories are one-hot encoded as vectors and the success of the models is compared with them as well. The reason for using one-hot encoding is to see whether the embedding vectors are more successful than one-hot encoders. Even if they derive from the same source, models with embedding features are more successful than one-hots.

Even if the classification results are mediocre because of the imbalanced data set and not having various features, the results show that, appending customer embedding vectors as features have improved the success of the classification model. As a result, customer representations created by embeddings improve the models. Fraud detection is examined for this study but customer embedding can be applied as features for other classification problems as well.